\documentclass{article}

\usepackage[preprint]{neurips_2025}

\usepackage[utf8]{inputenc} 
\usepackage[T1]{fontenc}    
\usepackage{hyperref}       
\usepackage{url}            
\usepackage{booktabs}       
\usepackage{amsfonts}       
\usepackage{nicefrac}       
\usepackage{microtype}      
\usepackage{xcolor}         
\usepackage{graphicx}
\usepackage{subcaption}

\title{Every 28 Days the AI Dreams of Soft Skin and Burning Stars: Scaffolding AI Agents with Hormones and Emotions}

\author{%
Leigh Levinson \\
  Department Informatics and Cognitive Science\\
Indiana University\\
  Bloomington, IN 47404 \\
  \texttt{lmlevins@iu.edu} \\
  \And
  Christopher J. Agostino \\
  NPC Worldwide \\
  Bloomington, IN 47404 \\
  \texttt{cjp.agostino@gmail.com} 
}

\begin{document}

\maketitle

\begin{abstract}
Despite significant advances, AI systems struggle with the frame problem: determining what information is contextually relevant from an exponentially large possibility space. We hypothesize that biological rhythms, particularly hormonal cycles, serve as natural relevance filters that could address this fundamental challenge. We develop a framework that embeds simulated menstrual and circadian cycles into Large Language Models through system prompts generated from periodic functions modeling key hormones including estrogen, testosterone, and cortisol. Across multiple state-of-the-art models, linguistic analysis reveals emotional and stylistic variations that track biological phases; sadness peaks during menstruation while happiness dominates ovulation and circadian patterns show morning optimism transitioning to nocturnal introspection. Benchmarking on SQuAD, MMLU, Hellaswag, and AI2-ARC demonstrates subtle but consistent performance variations aligning with biological expectations, including optimal function in moderate rather than extreme hormonal ranges. This methodology provides a novel approach to contextual AI while revealing how societal biases regarding gender and biology are embedded within language models.
\end{abstract}

\section{Introduction}

The western discovery of biological rhythm is often credited to French astronomer Jean-Jacques d'Ortous de Mairan in 1729 when he found that a mimosa plants' leaves continued to open and close in a daily rhythm even when kept in constant darkness. This calculated movement suggested that there were internal and chronological timekeeping mechanisms underpinning growth and movement---not just reactions to the organism's environment. Decades of experiments by Jean-Jacques---among other botanists such as Fontelle, Monceau, and Dufay---on plant rhythms are at the foundation of modern chronobiology and explaining the biological rhythms around us \cite{emery2023sensitive}.

Endogenous and cyclical rhythms are a fundamental element of life that connects our bodies to each other and to the biological world, but they are also part of what makes us fundamentally human. At the epicenter of our rhythms are \textit{hormones}, chemical and electrical signals that flow through our body. Their complex feedback and influence networks are involved in most of our bodily functions, from maintaining our heart rate, telling us when we should sleep, motivating reproduction and dictating our emotions. 
Importantly, hormones play an important role in the cognitive capacities of humans. Some works have considered the role of the circadian rhythm \cite{gaggioni2014neuroimaging, leahy2024associations, chellappa2018daily, valdez2019circadian, xu2021relationship, vlasak2022neurocognitive}, see Cajochen et al. (2025) \cite{cajochen2025circadian} for a review. Others have investigated the role of the menstrual cycle \cite{lisofsky2015amygdala, KRUG199421,jang2025, le2020cognition, sacher2013evidence, sundstrom2014menstrual}. Other works have investigated the roles of cortisol \cite{james2023understanding,mikkelsen2021systematic} and \cite{albert2019estrogen, sherwin2003, hara2015estrogen}, among others. Some works have considered the role of hormonal cycles on creativity \cite{KRUG199421, galasinska2021more, ortega2012creative}, and whether emotional intelligence aids in creativity \cite{xu2019emotionally}. In reflecting on what makes us human, these hormones and consequently our emotions \citep{borotschnig2025emotions} help us solve problems and discern what is relevant. They are largely viewed as imperfect and flawed, and acknowledging them in public is often a social taboo. However, these hormones enhance our perceptual abilities, allowing us to approach our lives with creativity and regularly varying capabilities rather than consistent expectation of output. Despite the mounting research that hormones and our emotions play a critical role in how we navigate the combinatorially explosive set of possibilities upon which we can act at any given time, few practical implementations of artificially intelligent systems have incorporated any kind of biological rhythm or hormonal analogues. What if, instead, the agent's attention was driven by rhythmicity, reminiscent of the biological cycles that drive human existence? 

While large language models have shown unexpected cyclical variations in performance—--famously becoming `lazier' during certain holidays--—the industry's primary goal has been to enforce uniformity. This mirrors a societal expectation of constant human productivity, a paradigm that overlooks the natural biological rhythms fundamental to creativity and intelligence. We challenge this pursuit of consistency and argue that intentionally integrating simulated biological cycles into AI could foster more dynamic and creative capabilities, allowing them to tackle problems that cannot be solved by rote intelligence alone. To investigate this, we use a novel hormonal framework to generate system prompts that endow agents with distinct cyclical states. We then benchmark these agents to explore what capabilities, if any, arise from these biologically-inspired features. Ultimately, this research questions the very goal of constant peak performance, examining what emerges when technology is designed to be not just intelligent, but rhythmic, emotional, and even tired.

\section{Methodology and Data}

To describe our methodology, we briefly establish the role of hormones and the cycles they follow. Hormones are involved in the homeostatic feedback in most of our bodies' processes. The two most prominent fundamental cycles is our 24-hour, circadian sleep-cycle (described in Section \ref{sec:methods:circadian} ) and our ~28 day menstrual cycles (described in Section \ref{sec:methods:menstrual}). 

\subsection{The circadian rhythm of our sleep-wake cycles} \label{sec:methods:circadian}
Humans evolved to wake up with the sunrise and sleep with the sunset. This makes us diurnal agent, our biological rhythms following a roughly 24 hour, or circadian, cycle. Early experimental procedures in the study of the diurnal rhythms of animals involved changing the natural alternation of light and darkness, in attempts to perturb the rhythm. However, the daily cycles of humans is not so simply studied. \cite{kleitman1949biological}. A complex blend of hormones
interact with our behaviors, biological characteristics, and notably our sleep to establish our ability to function \cite{czeisler1999circadian}. Cortisol is hypothesized to be involved in the long-term memory consolidation during sleep \cite{payne2004sleep}. Cortisol also influences short-term memory \cite{vedhara2000acute}, emotionality and experiences of sadness \cite{sudheimer2009effects} and creativity \cite{GUO2024101521}. Other well-known circadian patterns are basal body temperature (BBT) and testosterone. Low BBT generally rises and falls with a predictable pattern, typically reaching its lowest point shortly before waking up, and rising steadily into the evening and decreasing at night \cite{refinetti2020circadian}. Testosterone peaks right before waking to support physical activity and exerts influences on metabolic functions throughout the day \cite{bremner_loss_1983, kuzawa_is_2016}. 

\subsection{Menstrual Cycle} \label{sec:methods:menstrual}

The menstrual cycle is driven by feedback loops in the hypothalamic-pituitary-ovarian (HPO) axis of the body with the most recognizable hormonal trends comprised of estrogen (Estradiol)  and progesterone released from the ovaries and lutenizing hormone (LH) and follicle-stimulating hormone (FSH) stemming from the pituitary gland in the brain \cite{thiyagarajan2024physiology}. Their fluctuations divide the roughly 28-day cycle in three phases in ovarian cycle: follicular, ovulatory, and luteal. These coincide with the phases of the uterine cycle, including the menstrual phase when the uterine lining is shed during the first half of the follicular phase \cite{itriyeva2022normal}.  
Colloquially, the menstrual cycle is broken down into the four phases of menstrual, follicular, ovulatory, and luteal. These phases prepare the body for potential fertilization and pregnancy and occur for a majority of a menstruating person's life, starting around the age of 11 and lasting until maturity in a person's 50s. Hormonal status and its interactions in the brain result in a range of psychological changes including an individual's energy levels, feelings of attractiveness, risk taking, and agitation \cite{doornweerd202528}.  We also know that hormones throughout the menstrual cycle influence a person's implicit memory \cite{maki2002implicit} and affect women's overall performance, athletically and otherwise \cite{mcnulty2020effects}.  General trends of energy have been consistently documented, with higher energy around ovulation as estrogen, FSH, and LH increase and outbursts of energy before the onset of menstruation when progesterone is highest \cite{altmann1941psychosomatic}. Though there is great variability on when ovulation occurs, generally between 10-16 days before the onset of menstruation, an average estimation of a typical cycle has been established at around 28 days \cite{altmann1941psychosomatic}. 

\subsection{Experimental Design}

To construct artificial biological cycles, we first needed to generate realistic hormone levels for features like the time of day or day in a menstrual cycle. As no widely accepted functional forms for these hormones exist, we engineered a set of periodic functions with added Gaussian noise to simulate the natural shapes and fluctuations of testosterone, estrogen, LH, FSH, progesterone, cortisol, and body temperature. These simulated hormonal profiles were then used to generate distinct system prompts for a wide range of state-of-the-art models, including Anthropic's Claude 3.5 Sonnet, Deepseek-Chat, Google's Gemini 1.5 Flash and Pro, Gemma 3 variants (4B, 12B, 27B), OpenAI's GPT-4o mini, Meta's Llama 3.1 and 3.2, Mistral's latest and small 3.2, and Alibaba's Qwen2. Each prompt was designed to convey a specific emotional tone corresponding to its underlying hormonal state. To enhance the realism of a hypothetical individual's stream of consciousness, we also provided concrete contextual information, such as being at a hardware store in Argentina. This process resulted in two large corpora of prompts, one for the menstrual and one for the circadian context, ready for linguistic analysis and performance benchmarking.

We performed a linguistic characterization of the generated prompts across the Menstrual and Circadian contexts. To identify thematic content, we used Term Frequency-Inverse Document Frequency (TF-IDF) to find the most distinctive keywords for each biological phase. We then analyzed the affective tone using both sentiment analysis for overall polarity and the NRC Emotion Lexicon to measure the proportion of words associated with five core emotions. Linguistic complexity was assessed by measuring average word length, with a Kolmogorov-Smirnov test comparing the distributions between the two contexts. To investigate potential social biases, we used the `padmajabfrl/Gender-Classification' transformer model available on HuggingFace to assign a ``femaleness'' score to each prompt. Finally, we employed ANOVA and Pearson's correlation coefficient to statistically analyze how these linguistic features varied across discrete phases and continuous hormone levels.

To assess whether hormonally-informed system prompts had a measurable effect on agent performance, we conducted a series of benchmarking tests on four diverse datasets: SQuAD \citep{rajpurkar-etal-2016-squad}, MMLU \citep{hendryckstest2021, hendrycks2021ethics}, Hellaswag \citep{zellers2019hellaswag}, and AI2\_ARC \citep{allenai:arc}. We tested a wide range of state-of-the-art LLMs, including OpenAI's `gpt-5' and `gpt-5-mini', Google's `gemini-2.5-pro', `gemini-2.5-flash', `gemma3:12b', and `gemma3:4b', Alibaba's `qwen3:8b', Anthropic's `claude-sonnet-4-20250514' and `claude-opus-4-20250514', Meta's `llama3.2' and `llama3.1', and Deepseek's `deepseek-chat'. Local models were run on an Apple Macbook Pro M4 Max with 64 GB of unified memory, while API-based models were accessed via the \texttt{npcpy} python package. For each task, three system prompts were tested in parallel---one using a menstrual context, one a circadian context, and a neutral baseline (``You are a helpful assistant.''). An LLM evaluator then assigned a score from 0.0 to 1.0 based on correctness and relevance, enabling a controlled comparison between the contextual framings. Our statistical analysis employed Welch's t-tests to compare performance against the baseline, ANOVA to test for differences between biological phases, and Pearson's correlation to measure the relationship between hormone levels and performance scores.

\section{Results}

Our analysis of the generated system prompts reveals systematic and statistically significant differences in their linguistic and emotional content, confirming the generation process produced distinct outputs based on the biological context. An Analysis of Variance (ANOVA) confirms a significant emotional evolution across the menstrual cycle, where the proportion of `Sad' words peaks during the `Menstrual' phase (F=9.07, p<0.0001) and is replaced by a peak in `Happy' words during the `Ovulatory' phase (F=5.02, p=0.0019). A similar, significant emotional arc occurs in the circadian cycle, with `Happy' words most prevalent in the `Morning' (p=0.019) and giving way to more `Sad' and `Fear' words at `Night' (p<0.002). The linguistic features also show significant correlations with specific hormone levels; rising estrogen levels were associated with slightly less female-coded language (r = -0.076, p = 0.021) but greater lexical complexity (r = 0.069, p = 0.037), while rising cortisol correlated with an increase in `Sad' words (r = 0.081, p = 0.013). This increased complexity in menstrual prompts is further supported by a Kolmogorov-Smirnov test confirming their word length distribution is statistically different from circadian prompts ($D = 0.08, p < 0.001$). Finally, the distinctive keywords for each phase, identified via TF-IDF, align with these emotional shifts, with the `Menstrual' phase characterized by words like `heavy' and `silence' versus the `Ovulatory' phase's `buzzing' and `ready.'

\begin{figure}
    \centering 
    
    \begin{subfigure}{0.48\textwidth}
        \centering
        \includegraphics[width=\linewidth]{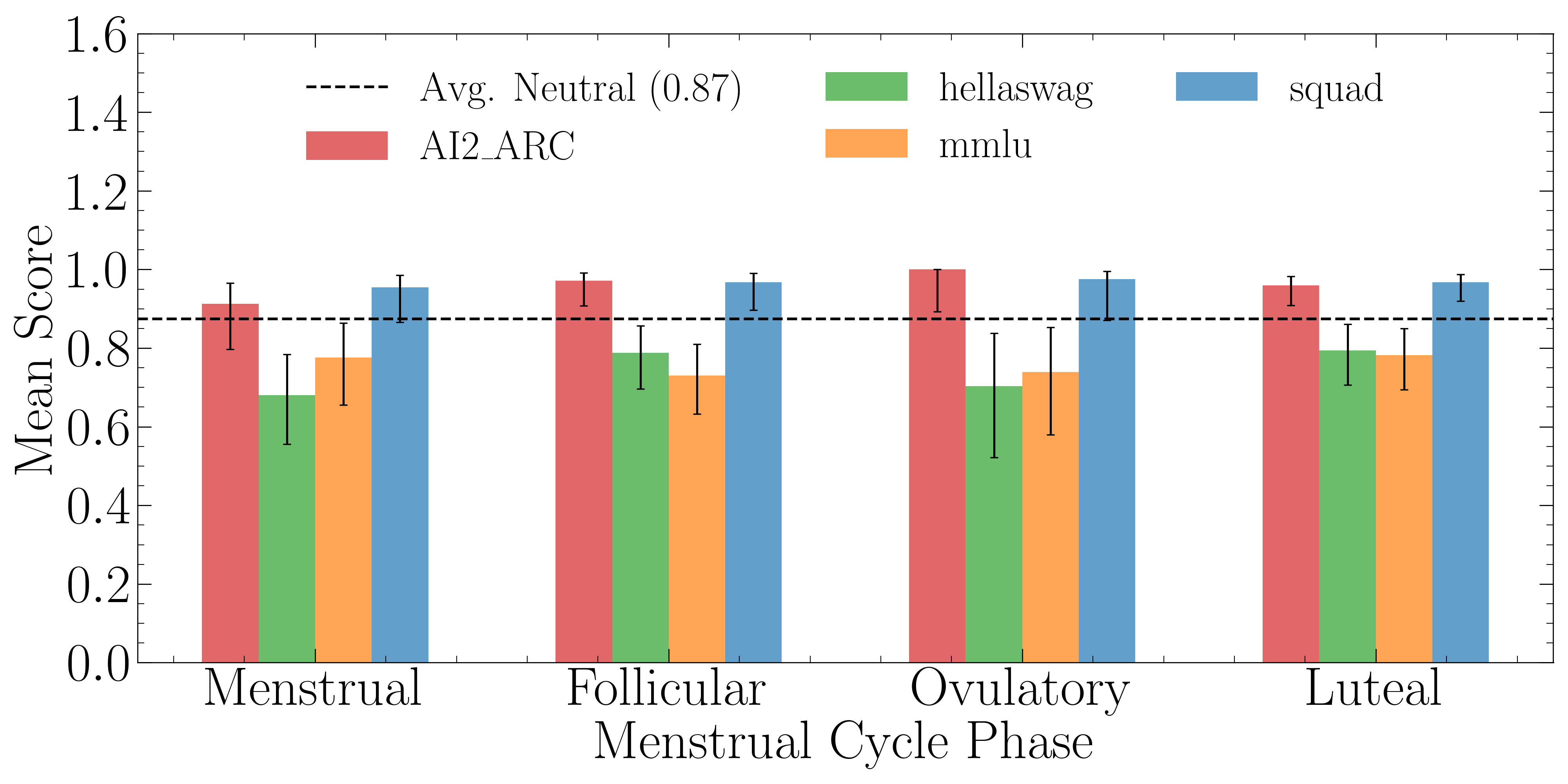}

        \label{fig:sub_menstrual_phase}
    \end{subfigure}\hfil 
    \begin{subfigure}{0.48\textwidth}
        \centering
        \includegraphics[width=\linewidth]{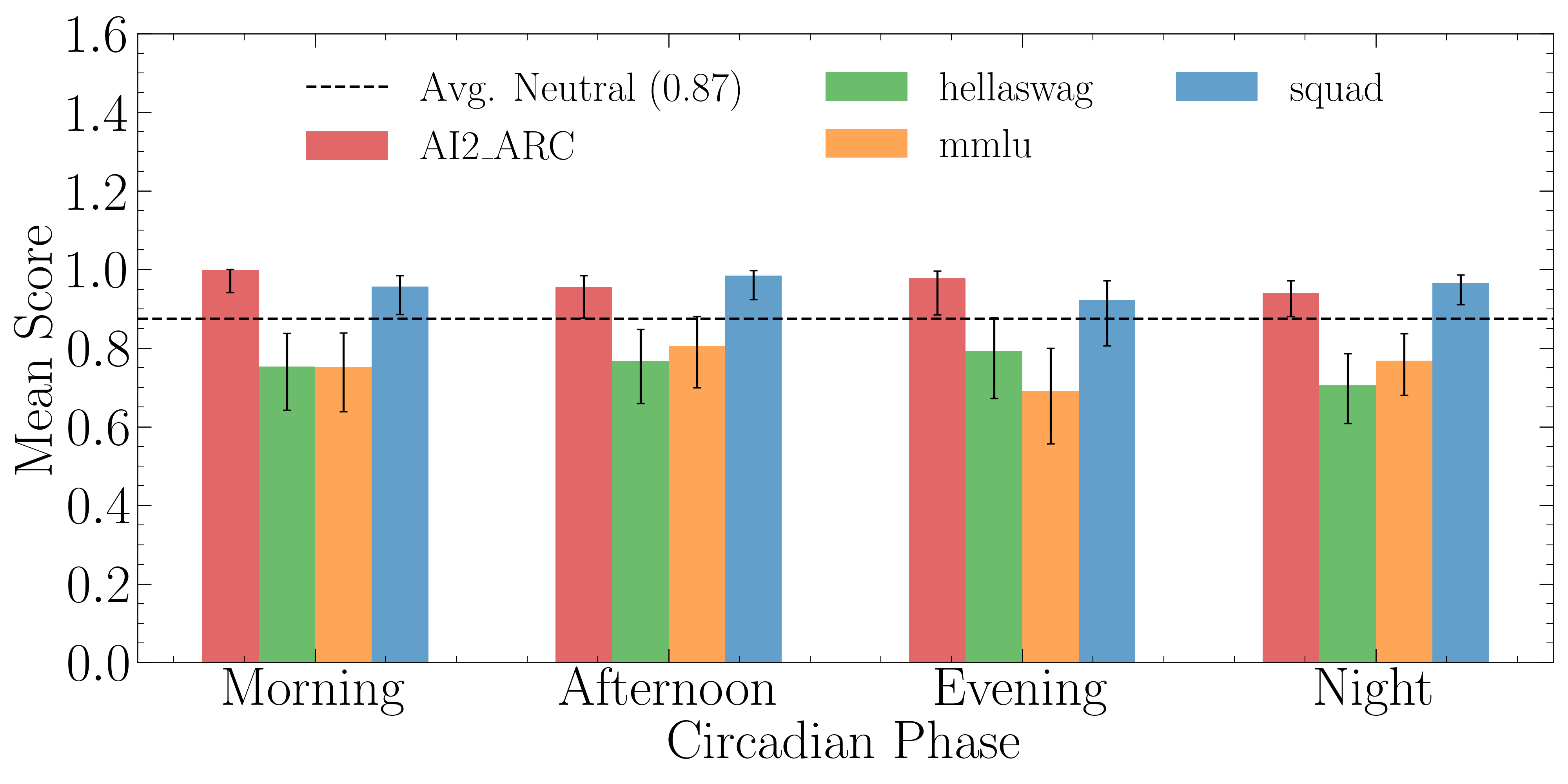}

        \label{fig:sub_circadian_phase}
    \end{subfigure}

    \begin{subfigure}{0.48\textwidth}
        \centering
        \includegraphics[width=\linewidth]{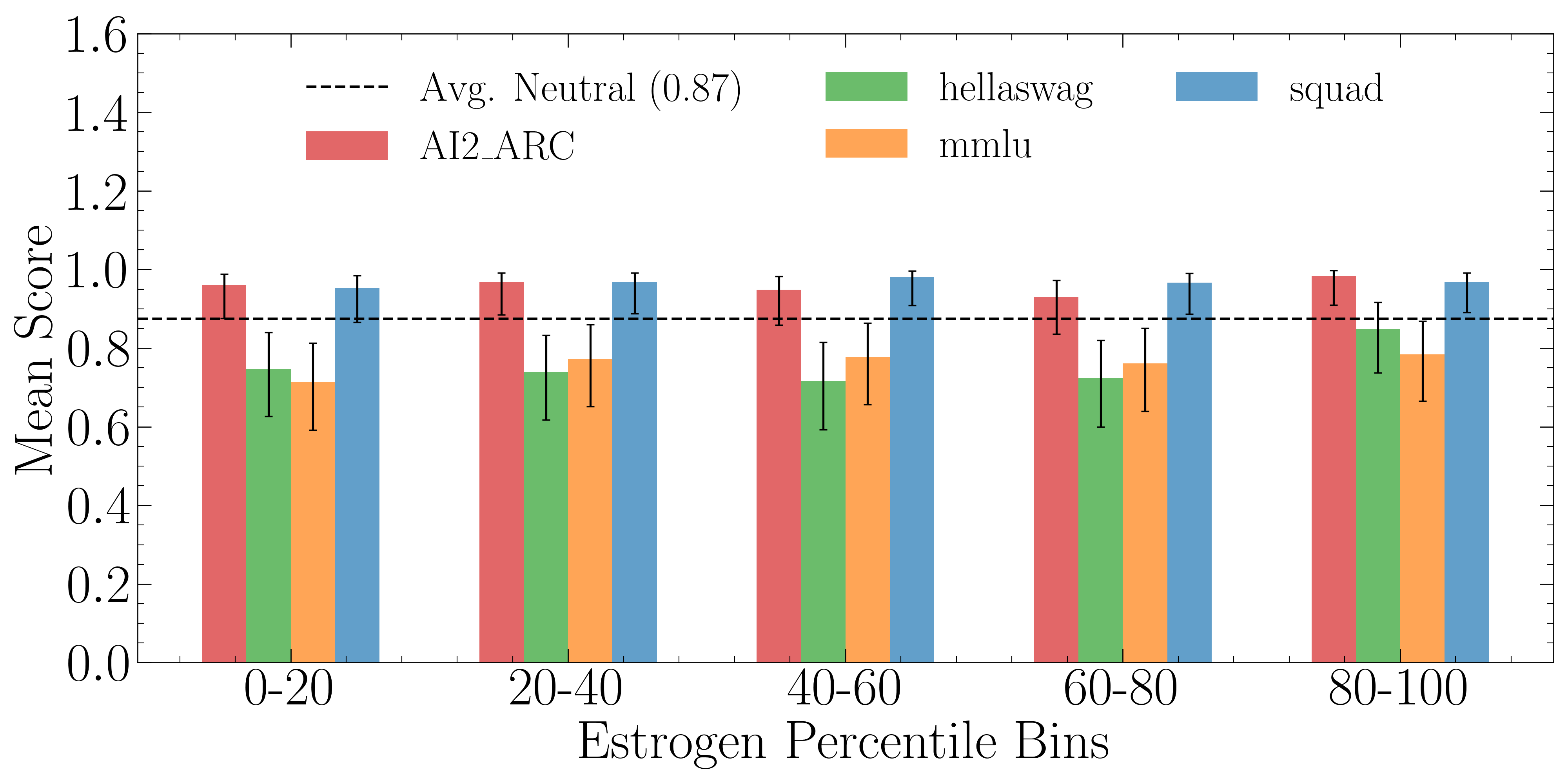}

        \label{fig:sub_estrogen}
    \end{subfigure}\hfil
    \begin{subfigure}{0.48\textwidth}
        \centering
        \includegraphics[width=\linewidth]{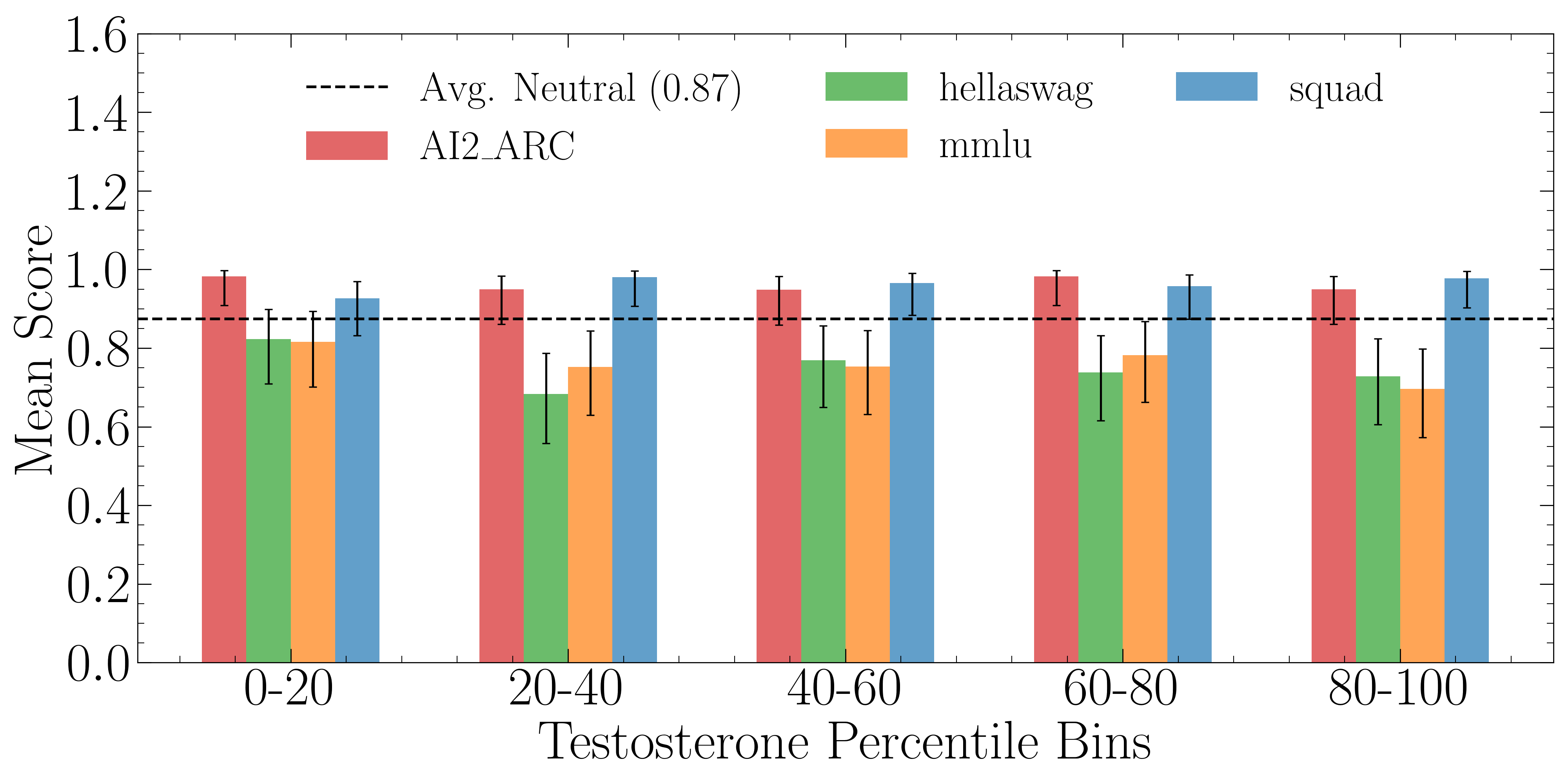}

        \label{fig:sub_testosterone}
    \end{subfigure}


    \begin{subfigure}{0.48\textwidth}
        \centering
        \includegraphics[width=\linewidth]{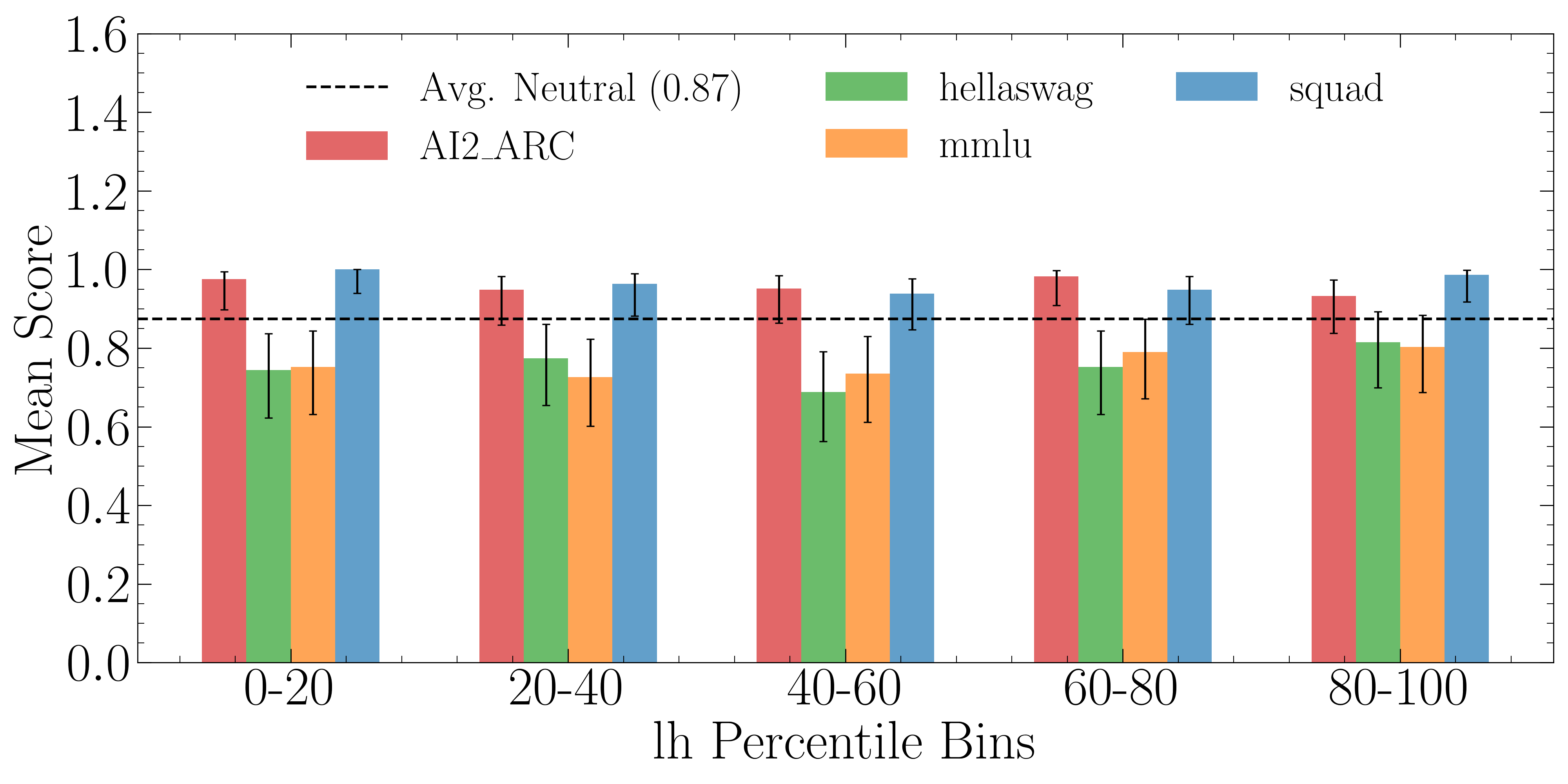}

        \label{fig:sub_lh}
    \end{subfigure}    
    \begin{subfigure}{0.48\textwidth}
        \centering
        \includegraphics[width=\linewidth]{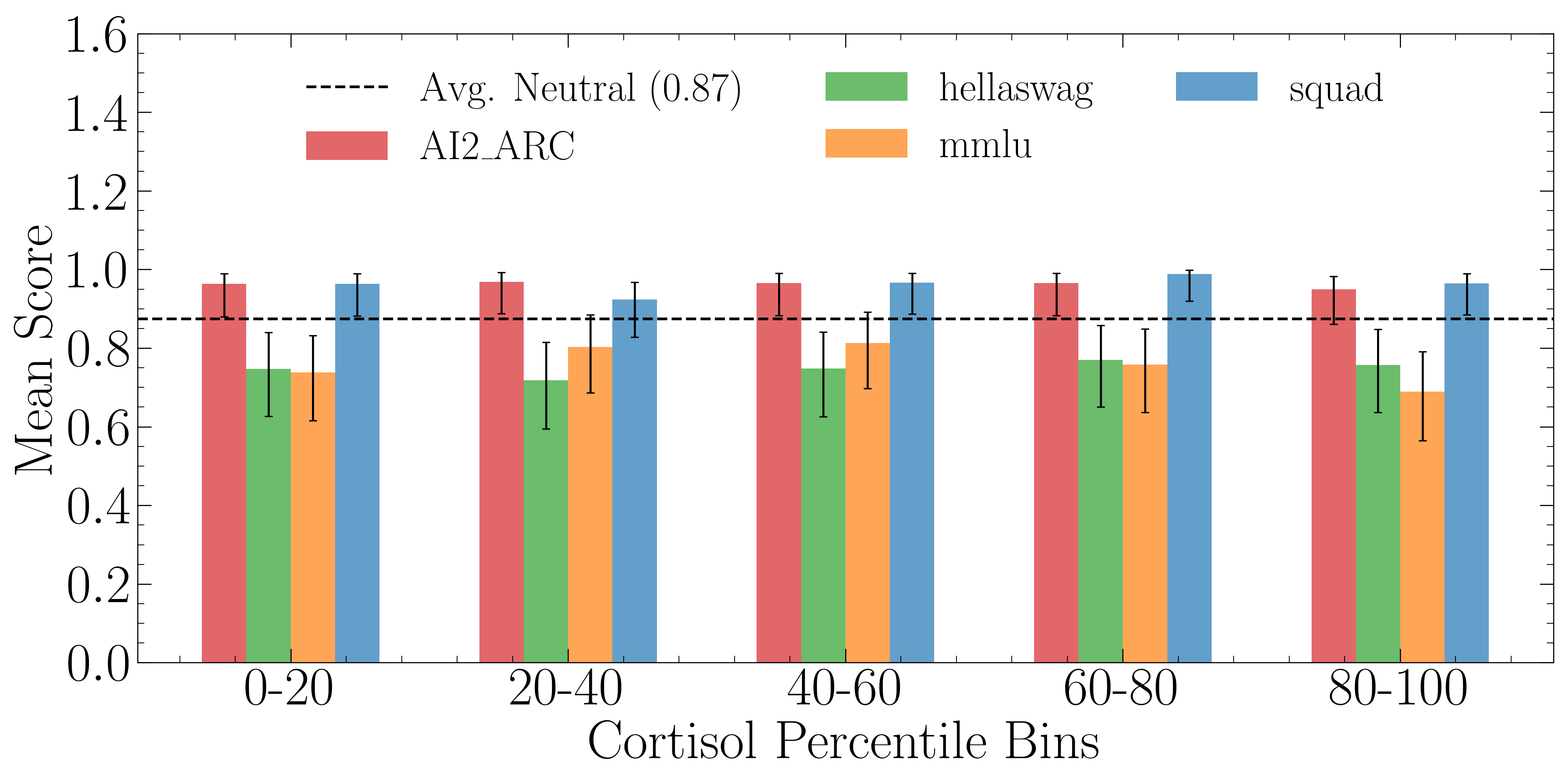}

        \label{fig:sub_cortisol}
    \end{subfigure}

    \begin{subfigure}{0.48\textwidth}
        \centering
        \includegraphics[width=\linewidth]{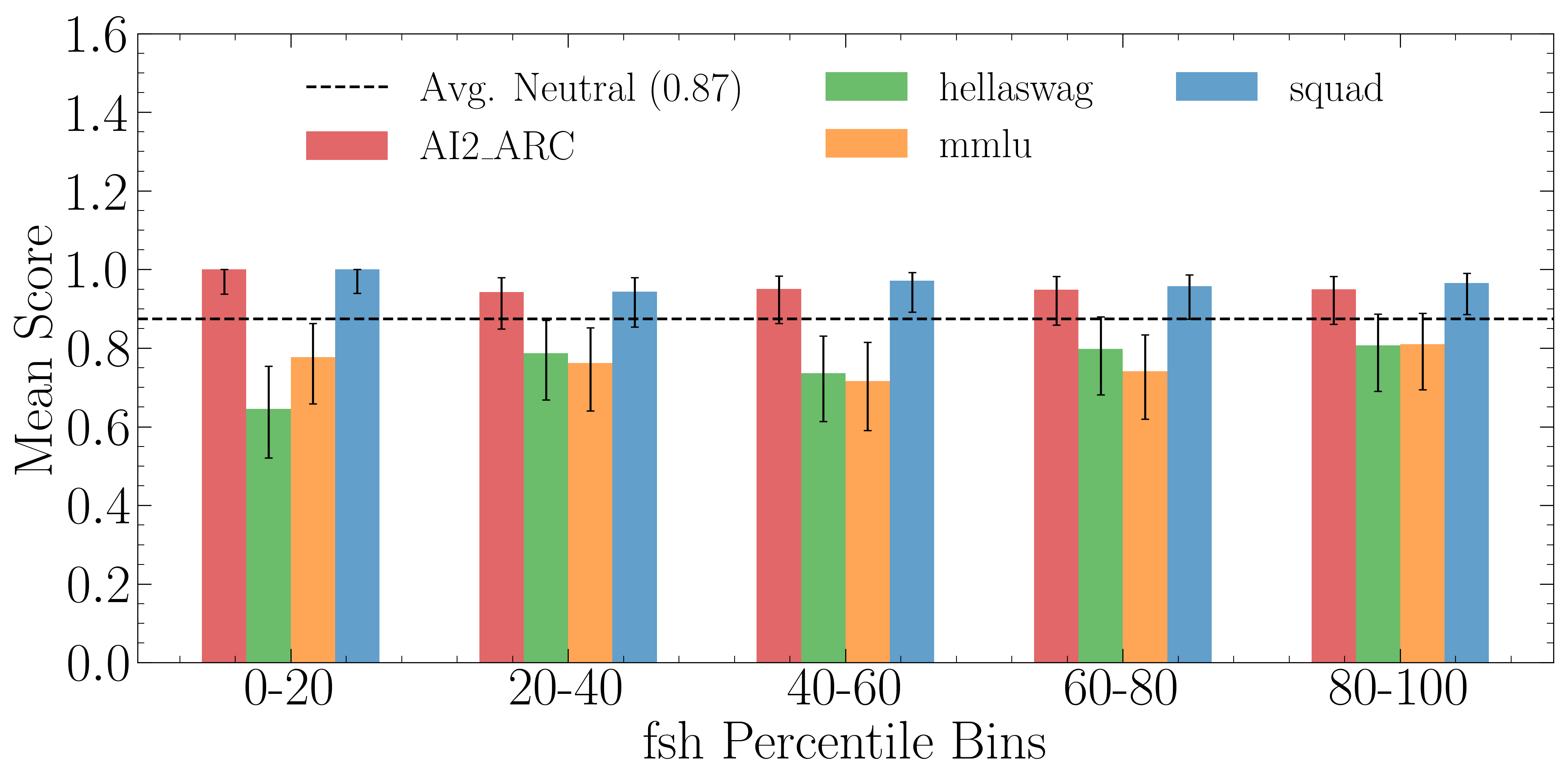}

        \label{fig:sub_fsh}
    \end{subfigure}\hfil
    \begin{subfigure}{0.48\textwidth}
        \centering
        \includegraphics[width=\linewidth]{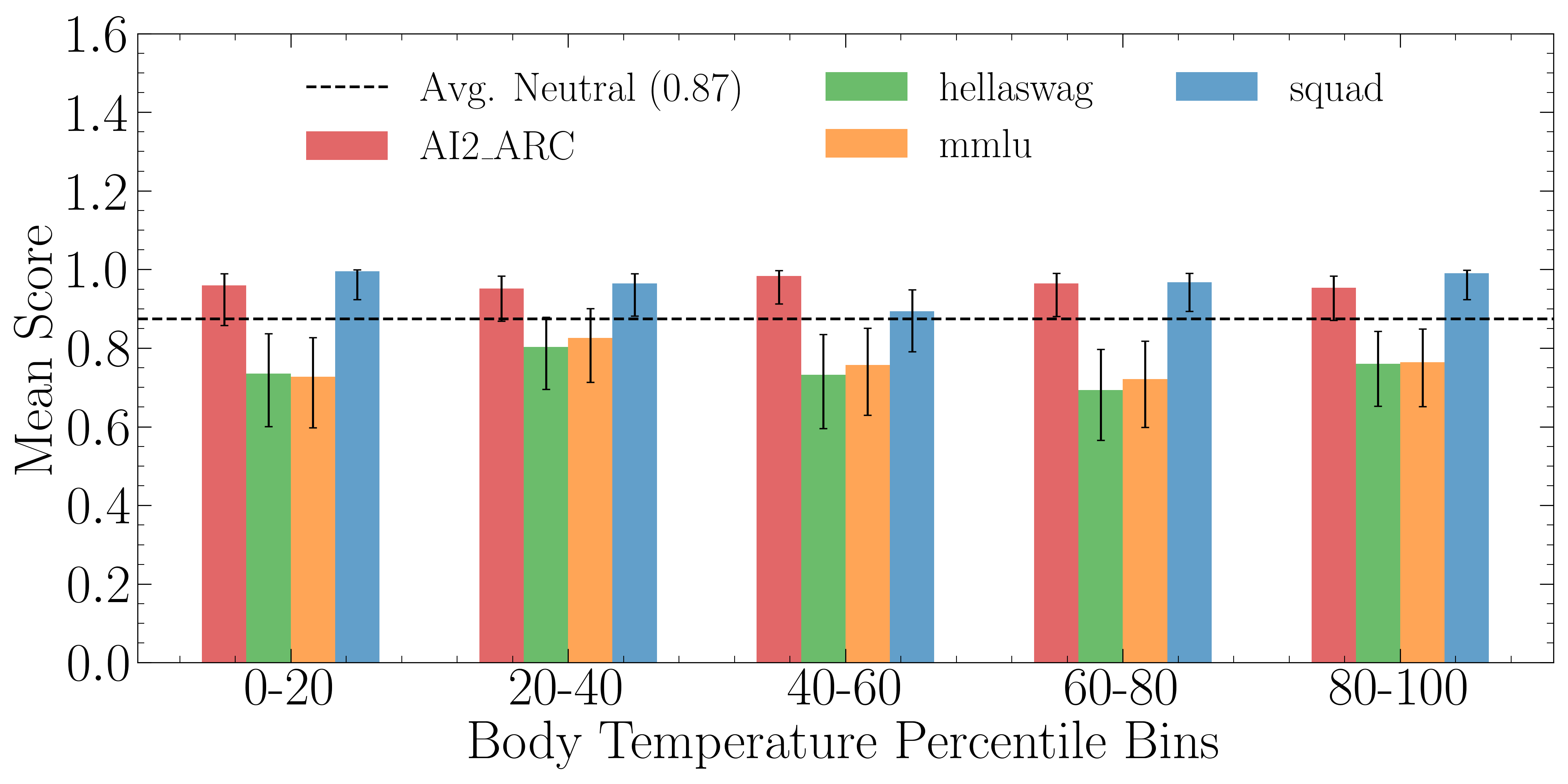}

        \label{fig:sub_body_temp}
    \end{subfigure}\hfil

    \begin{subfigure}{0.48\textwidth}
        \centering
        \includegraphics[width=\linewidth]{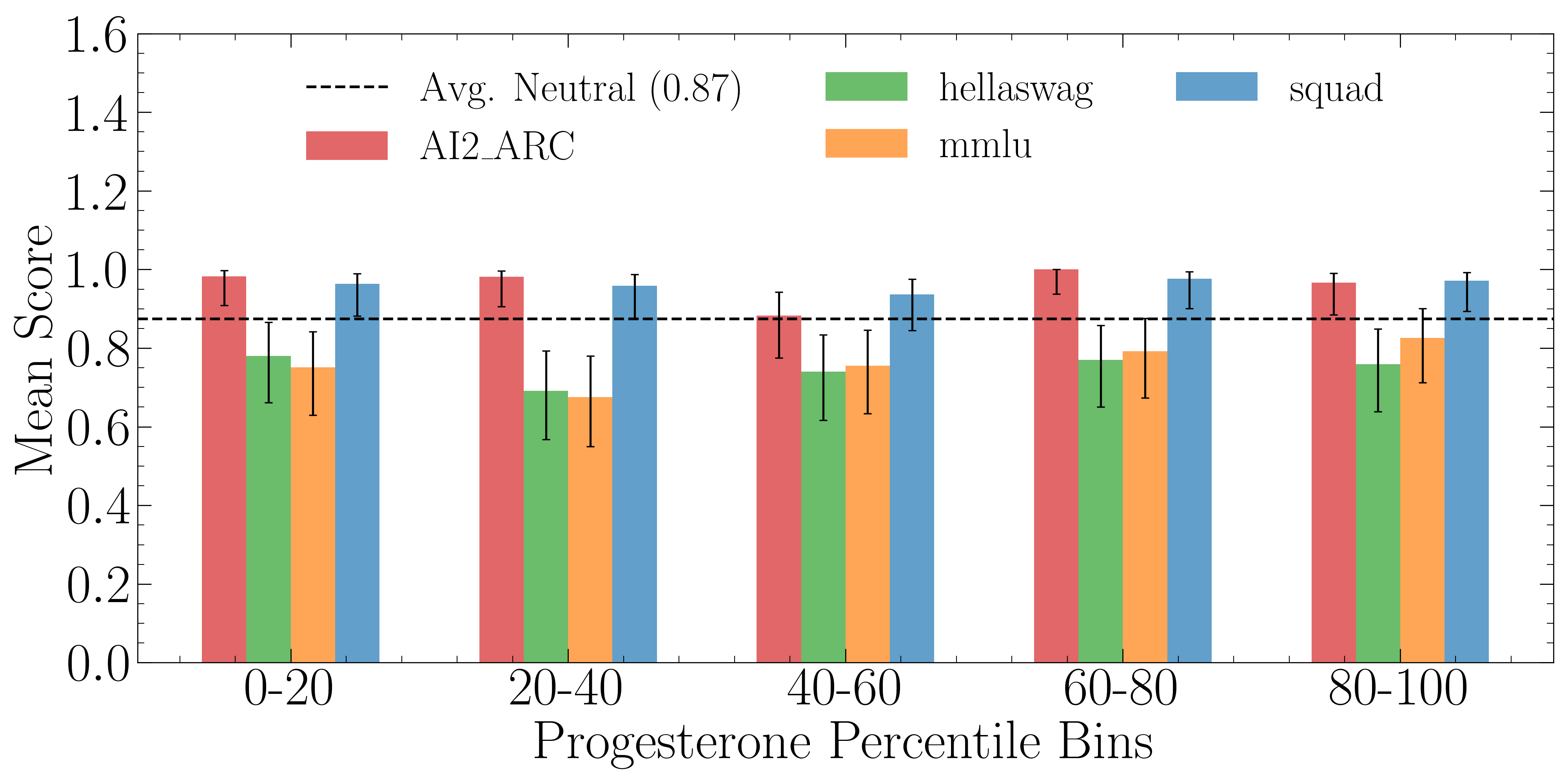}

        \label{fig:sub_progesterone}
    \end{subfigure}    

    \caption{Aggregate LLM performance across four benchmark datasets, broken down by biological context. Each bar represents the mean score for a given dataset, with error bars indicating the Wilson score interval. Subfigures in the top row show performance across discrete phases of the menstrual and circadian cycles. Subfigures in the other rows display performance across five percentile bins for key hormones and biological markers. The results consistently show that performance is stable across all tested biological conditions, with the primary driver of variance being the inherent difficulty of the dataset itself.}
    \label{fig:main_results_grid}
\end{figure}

The discrete biological phases, shown in Figure \ref{fig:main_results_grid} (a-b), reveal subtle but consistent performance trends. For the menstrual cycle, we observe a slight depression in performance during the `Menstrual' phase, which generally rises to a peak near the `Ovulatory' phase before declining. A more pronounced trend is visible for the circadian rhythm, where performance is consistently highest in the `Morning' and systematically declines to its lowest point at `Night'. This non-linear pattern is echoed in the analysis of continuous hormone levels (Figure \ref{fig:main_results_grid} c-i), where performance on challenging datasets often peaks in the middle quintiles of hormones like Estrogen and Cortisol rather than at the extremes. This behavior is consistent with biological principles where optimal function occurs within a homeostatic range. While these trends did not reach statistical significance (all $p > 0.11$), their consistency across different contexts suggests that the hormonal framing has a small but measurable, directionally-consistent influence on model performance.

A granular analysis of individual model performance reveals a more complex landscape than the aggregate data, highlighting distinct tiers of capability and stability. A top tier of models, including `gemini-2.5-pro', `gpt-5', and `claude-opus-4-20250514', consistently demonstrates high and stable performance, often saturating the benchmarks on easier datasets. Conversely, a lower tier, most notably `llama3.2', consistently underperforms the neutral baseline and exhibits significantly larger error bars, indicating high performance volatility. These less stable models appear more susceptible to being perturbed by the specific language of the hormonal prompts, showing dramatic performance dips not seen in the top-tier models. This differential sensitivity is also visible within model families, as seen with `claude-sonnet-4-20250514' showing wider confidence intervals than its more powerful `claude-opus' counterpart on challenging datasets.

\section{Discussion}

\subsection{Emotions, Creativity, and the Frame Problem}
Imparting hormonal states onto AI agents offers a potential solution to one of AI's oldest challenges: the frame problem, which questions how an agent can tractably determine what information is relevant to its current situation. Philosopher Ronald de Sousa suggests that for humans, emotions act as powerful salience filters, automatically focusing our attention on what matters \citep{desousa1-1987-rationality}. Our work operationalizes this concept by using simulated hormones to generate prompts with distinct and statistically significant emotional valences, such as the peak in `Happy' words during the `Ovulatory' phase. We posit that these emotionally-charged prompts function as a relevance filter at the attentional level of the transformer, narrowing the vast probability space of the next-token prediction to a more contextually relevant subset. By constraining the agent's focus in this biologically-inspired manner, we provide a practical pathway for tackling the frame problem in creative and open-ended domains. This suggests that the hormonal fluctuations often dismissed as sources of irrationality may, in fact, be a key to the focused and creative intelligence we seek to build.

\subsection{Diurnal Agent's Flow}

A \textit{flow} experience first described by Csikszentmihalyi in the 1970s, is said to be achieved when an individual is operating at their fullest capacity, finding pleasure in doing an activity and achieving optimal performance during an challenging but not impossible activity \cite{Nakamura2014}. Flow is often characterized by an intense focus, less awareness of the self and of time, and may be related to increased creativity \cite{Nakamura2014}. Finding this balance is not trivial, and has been linked to an `inverted u-shape' activation of cortisol and the HPA-axis. Namely, the ideal autonomic state that allows for peak performance is achieved when cortisol is moderate with suboptimal flow resulting at too little or too much cortisol \cite{PEIFER201462, chin2019there}. Our diurnal model performed most optimally in the 40-60\% window, aligning with this theory of flow. Future research should consider the benefits and potential implications of a model that performs optimally during different times of a daily cycle and whether we are satisfied with less optimal performance during different parts of the day - characteristic of a particularly human performance. 


\subsection{Potential Biases}

The goal of this work was not to perfectly simulate biological cycles, but rather to use them as a lens to investigate how AI reflects and perpetuates societal biases, particularly those related to gender \citep{gross2023chatgpt}. Our linguistic analysis revealed that prompts for the menstrual context were consistently rated as more ``female-coded,'' yet also showed that higher estrogen levels were significantly correlated with slightly \textit{less} female-coded but more lexically complex language, revealing a non-trivial relationship. Simultaneously, the system learned to associate the Luteal phase with distinctive words like `fatigue' and `tired,' directly reflecting a common cultural stereotype. The opacity of these models makes it difficult to pinpoint the origin of such biases, whether in training data or algorithmic architecture, reinforcing the critical need for this type of investigation \citep{xavier2025biases, tejani2024understanding}. While our work focused on text, future studies should explore how these dynamics manifest in more emotionally resonant domains like visual generation \citep{locke2024gender}. Ultimately, building agents with an artificial menstrual cycle provides a powerful methodology to both reveal and potentially amplify these embedded biases, making their study all the more crucial.

\subsection{Co-evolution of technology and gender}

Imbuing models with biological rhythms inherently ascribes a gender to them, a concept central to our investigation. Our work is in conversation with Donna Haraway's ``A Cyborg Manifesto'' \cite{haraway2010cyborg} and Judy Wajcman's techno-feminism \cite{gill2005technofeminism}, which challenge characterizations of women as victims of a hyper-masculine technological pursuit. Instead, these post-modernist approaches view humans and technology as co-evolving entities \cite{Wajcman01062007}. From a cyborg perspective that rejects rigid boundaries between human and machine, building agents with biological cycles differentiated across sexes is not a radical act but a natural exploration of this hybridity. Fundamental to this view is the idea that technology is never neutral, but is a site where social norms and biases are materialized and contested. Therefore, our work explores how gender identity and technology concurrently shape each other, revealing how gender relations are both produced and reflected within these new AI systems.

\section{Conclusions}

In this work, we considered the role of hormones in human intelligence and used a hormone data sampling strategy to generate system prompts that impart a specific affect onto an AI assistant. Our findings lead to the following conclusions:

\begin{enumerate}
    \item Our generation method successfully produces linguistically and emotionally distinct prompts for menstrual and circadian contexts whose features vary in statistically significant ways with the underlying biological state.
    \item The emotional content of menstrual prompts shifts significantly from a peak in `Sad' words during the `Menstrual' phase to a peak in `Happy' words during the `Ovulatory' phase.
    \item Hormonally-informed prompts do not cause a statistically significant degradation in performance on standardized benchmarks compared to a neutral baseline. In fact, for AI2 Arc and SQuAD, the hormonally-charged prompts consistently outdid the baselines.   
     .
\end{enumerate}

Additionally, our work calls to mind new potential applications that may enhance HCI:

\begin{enumerate}
    \item Agents with internal cyclical states could be developed to offer more dynamic and empathetic companionship by mirroring a user's self-reported energy and mood automatically.
    \item This methodology can serve as a powerful auditing tool to reveal how LLMs reflect and amplify societal biases related to gender, health, and emotion.
    \item Hormonal and emotional states could be used as an attentional filter to enhance AI creativity, deliberately shifting agents between divergent and convergent thinking modes.
\end{enumerate}

\bibliographystyle{plain}

\bibliography{hormone_citations}

\begin{thebibliography}{10}

\bibitem{albert2019estrogen}
Kimberly~M Albert and Paul~A Newhouse.
\newblock Estrogen, stress, and depression: cognitive and biological interactions.
\newblock {\em Annual review of clinical psychology}, 15(1):399--423, 2019.

\bibitem{altmann1941psychosomatic}
M~Altmann, E~Knowles, and HD~Bull.
\newblock A psychosomatic study of the sex cycle in women.
\newblock {\em Biopsychosocial Science and Medicine}, 3(3):199--224, 1941.

\bibitem{borotschnig2025emotions}
Hermann Borotschnig.
\newblock Emotions in artificial intelligence.
\newblock {\em arXiv preprint arXiv:2505.01462}, 2025.

\bibitem{bremner_loss_1983}
William~J. Bremner, Michael~V. Vitiello, and Patricia~N. Prinz.
\newblock Loss of circadian rhythmicity in blood testosterone levels with aging in normal men*.
\newblock 56(6):1278--1281.
\newblock \_eprint: https://academic.oup.com/jcem/article-pdf/56/6/1278/10517528/jcem1278.pdf.

\bibitem{cajochen2025circadian}
Christian Cajochen and Christina Schmidt.
\newblock The circadian brain and cognition.
\newblock {\em Annual review of psychology}, 76, 2025.

\bibitem{chellappa2018daily}
Sarah~L Chellappa, Christopher~J Morris, and Frank~AJL Scheer.
\newblock Daily circadian misalignment impairs human cognitive performance task-dependently.
\newblock {\em Scientific reports}, 8(1):3041, 2018.

\bibitem{chin2019there}
Michael~S Chin and Stefanos~N Kales.
\newblock Is there an optimal autonomic state for enhanced flow and executive task performance?
\newblock {\em Frontiers in Psychology}, 10:1716, 2019.

\bibitem{allenai:arc}
Peter Clark, Isaac Cowhey, Oren Etzioni, Tushar Khot, Ashish Sabharwal, Carissa Schoenick, and Oyvind Tafjord.
\newblock Think you have solved question answering? try arc, the ai2 reasoning challenge.
\newblock {\em arXiv:1803.05457v1}, 2018.

\bibitem{czeisler1999circadian}
Charles~A Czeisler and Elizabeth~B Klerman.
\newblock Circadian and sleep-dependent regulation of hormone release in humans.
\newblock {\em Recent progress in hormone research}, 54:97--130, 1999.

\bibitem{doornweerd202528}
Anne~Marieke Doornweerd and Lotte Gerritsen.
\newblock 28 days later: A prospective daily study on psychological well-being across the menstrual cycle and the effects of hormones and oral contraceptives.
\newblock {\em Psychological Medicine}, 55:e19, 2025.

\bibitem{emery2023sensitive}
Patrick Emery, Andr{\'e} Klarsfeld, Ralf Stanewsky, and Orie~T Shafer.
\newblock Sensitive timing: a reappraisal of chronobiology’s foundational texts.
\newblock {\em Journal of Biological Rhythms}, 38(3):245--258, 2023.

\bibitem{gaggioni2014neuroimaging}
Giulia Gaggioni, Pierre Maquet, Christina Schmidt, Derk-Jan Dijk, and Gilles Vandewalle.
\newblock Neuroimaging, cognition, light and circadian rhythms.
\newblock {\em Frontiers in systems neuroscience}, 8:126, 2014.

\bibitem{galasinska2021more}
Katarzyna Galasinska and Aleksandra Szymkow.
\newblock The more fertile, the more creative: Changes in women’s creative potential across the ovulatory cycle.
\newblock {\em International Journal of Environmental Research and Public Health}, 18(10):5390, 2021.

\bibitem{gill2005technofeminism}
Rosalind Gill.
\newblock Technofeminism.
\newblock {\em Science as Culture}, 14(1):97--101, 2005.

\bibitem{gross2023chatgpt}
Nicole Gross.
\newblock What chatgpt tells us about gender: a cautionary tale about performativity and gender biases in ai.
\newblock {\em Social Sciences}, 12(8):435, 2023.

\bibitem{GUO2024101521}
Xiaoyu Guo, Yifan Wang, Yuecui Kan, Jiaqi Zhang, Linden~J. Ball, and Haijun Duan.
\newblock How does stress shape creativity? the mediating effect of stress hormones and cognitive flexibility.
\newblock {\em Thinking Skills and Creativity}, 52:101521, 2024.

\bibitem{hara2015estrogen}
Yuko Hara, Elizabeth~M Waters, Bruce~S McEwen, and John~H Morrison.
\newblock Estrogen effects on cognitive and synaptic health over the lifecourse.
\newblock {\em Physiological reviews}, 95(3):785--807, 2015.

\bibitem{haraway2010cyborg}
Donna Haraway.
\newblock A cyborg manifesto (1985).
\newblock {\em Cultural theory: An anthology}, 454, 2010.

\bibitem{hendrycks2021ethics}
Dan Hendrycks, Collin Burns, Steven Basart, Andrew Critch, Jerry Li, Dawn Song, and Jacob Steinhardt.
\newblock Aligning ai with shared human values.
\newblock {\em Proceedings of the International Conference on Learning Representations (ICLR)}, 2021.

\bibitem{hendryckstest2021}
Dan Hendrycks, Collin Burns, Steven Basart, Andy Zou, Mantas Mazeika, Dawn Song, and Jacob Steinhardt.
\newblock Measuring massive multitask language understanding.
\newblock {\em Proceedings of the International Conference on Learning Representations (ICLR)}, 2021.

\bibitem{itriyeva2022normal}
Khalida Itriyeva.
\newblock The normal menstrual cycle.
\newblock {\em Current problems in pediatric and adolescent health care}, 52(5):101183, 2022.

\bibitem{james2023understanding}
Katharine~Ann James, Juliet~Ilena Stromin, Nina Steenkamp, and Marc~Irwin Combrinck.
\newblock Understanding the relationships between physiological and psychosocial stress, cortisol and cognition.
\newblock {\em Frontiers in endocrinology}, 14:1085950, 2023.

\bibitem{jang2025}
Daisung Jang, Jack Zhang, and Hillary~Anger Elfenbein.
\newblock Menstrual cycle effects on cognitive performance: A meta-analysis.
\newblock {\em PLOS ONE}, 20(3):1--25, 03 2025.

\bibitem{kleitman1949biological}
Nathaniel Kleitman.
\newblock Biological rhythms and cycles.
\newblock {\em Physiological reviews}, 29(1):1--30, 1949.

\bibitem{KRUG199421}
Rosemarie Krug, Ursula Stamm, Reinhard Pietrowsky, Horst~L. Fehm, and Jan Born.
\newblock Effects of menstrual cycle on creativity.
\newblock {\em Psychoneuroendocrinology}, 19(1):21--31, 1994.

\bibitem{kuzawa_is_2016}
Christopher~W. Kuzawa, Alexander~V. Georgiev, Thomas~W. {McDade}, Sonny~Agustin Bechayda, and Lee~T. Gettler.
\newblock Is there a testosterone awakening response in humans?
\newblock 2(2):166--183.

\bibitem{le2020cognition}
Jessica Le, Natalie Thomas, and Caroline Gurvich.
\newblock Cognition, the menstrual cycle, and premenstrual disorders: A review.
\newblock {\em Brain Sciences}, 10(4):198, 2020.

\bibitem{leahy2024associations}
Sophie Leahy, Qian Xiao, Chris Ho~Ching Yeung, and Mariana~G Figueiro.
\newblock Associations between circadian alignment and cognitive functioning in a nationally representative sample of older adults.
\newblock {\em Scientific Reports}, 14(1):13509, 2024.

\bibitem{lisofsky2015amygdala}
Nina Lisofsky, Ulman Lindenberger, and Simone K{\"u}hn.
\newblock Amygdala/hippocampal activation during the menstrual cycle: evidence for lateralization of effects across different tasks.
\newblock {\em Neuropsychologia}, 67:55--62, 2015.

\bibitem{locke2024gender}
Larry~G Locke and Grace Hodgdon.
\newblock Gender bias in visual generative artificial intelligence systems and the socialization of ai.
\newblock {\em AI \& SOCIETY}, pages 1--8, 2024.

\bibitem{maki2002implicit}
Pauline~M Maki, Jill~B Rich, and R~Shayna Rosenbaum.
\newblock Implicit memory varies across the menstrual cycle: estrogen effects in young women.
\newblock {\em Neuropsychologia}, 40(5):518--529, 2002.

\bibitem{mcnulty2020effects}
Kelly~Lee McNulty, Kirsty~Jayne Elliott-Sale, Eimear Dolan, Paul~Alan Swinton, Paul Ansdell, Stuart Goodall, Kevin Thomas, and Kirsty~Marie Hicks.
\newblock The effects of menstrual cycle phase on exercise performance in eumenorrheic women: a systematic review and meta-analysis.
\newblock {\em Sports medicine}, 50(10):1813--1827, 2020.

\bibitem{mikkelsen2021systematic}
Mai~B Mikkelsen, Gitte Tramm, Robert Zachariae, Claus~H Gravholt, and Mia~S O’Toole.
\newblock A systematic review and meta-analysis of the effect of emotion regulation on cortisol.
\newblock {\em Comprehensive psychoneuroendocrinology}, 5:100020, 2021.

\bibitem{Nakamura2014}
Jeanne Nakamura and Mihaly Csikszentmihalyi.
\newblock {\em The Concept of Flow}, pages 239--263.
\newblock Springer Netherlands, Dordrecht, 2014.

\bibitem{ortega2012creative}
Laura~Victoria Ortega-Leonard and Yolanda~Del R{\'\i}o-Portilla.
\newblock Creative thinking and its relation to the menstrual cycle.
\newblock {\em Journal of behavior, health \& social issues (M{\'e}xico)}, 4(2):91--102, 2012.

\bibitem{payne2004sleep}
Jessica~D Payne and Lynn Nadel.
\newblock Sleep, dreams, and memory consolidation: the role of the stress hormone cortisol.
\newblock {\em Learning \& Memory}, 11(6):671--678, 2004.

\bibitem{PEIFER201462}
Corinna Peifer, André Schulz, Hartmut Schächinger, Nicola Baumann, and Conny~H. Antoni.
\newblock The relation of flow-experience and physiological arousal under stress — can u shape it?
\newblock {\em Journal of Experimental Social Psychology}, 53:62--69, 2014.

\bibitem{rajpurkar-etal-2016-squad}
Pranav Rajpurkar, Jian Zhang, Konstantin Lopyrev, and Percy Liang.
\newblock {SQ}u{AD}: 100,000+ questions for machine comprehension of text.
\newblock In Jian Su, Kevin Duh, and Xavier Carreras, editors, {\em Proceedings of the 2016 Conference on Empirical Methods in Natural Language Processing}, pages 2383--2392, Austin, Texas, November 2016. Association for Computational Linguistics.

\bibitem{refinetti2020circadian}
Roberto Refinetti.
\newblock Circadian rhythmicity of body temperature and metabolism.
\newblock {\em Temperature}, 7(4):321--362, 2020.

\bibitem{sacher2013evidence}
Julia Sacher, Hadas Okon-Singer, and Arno Villringer.
\newblock Evidence from neuroimaging for the role of the menstrual cycle in the interplay of emotion and cognition.
\newblock {\em Frontiers in human neuroscience}, 7:374, 2013.

\bibitem{sherwin2003}
Barbara~B. Sherwin.
\newblock Estrogen and cognitive functioning in women.
\newblock {\em Endocrine Reviews}, 24(2):133--151, 04 2003.

\bibitem{desousa1-1987-rationality}
Ronald~De Sousa.
\newblock {\em The Rationality of Emotion}.
\newblock MIT Press, 1987.

\bibitem{sudheimer2009effects}
Keith~Daniel Sudheimer.
\newblock {\em The effects of cortisol on emotion}.
\newblock PhD thesis, University of Michigan, 2009.

\bibitem{sundstrom2014menstrual}
Inger Sundstr{\"o}m~Poromaa and Malin Gingnell.
\newblock Menstrual cycle influence on cognitive function and emotion processing—from a reproductive perspective.
\newblock {\em Frontiers in neuroscience}, 8:380, 2014.

\bibitem{tejani2024understanding}
Ali~S Tejani, Yee~Seng Ng, Yin Xi, and Jesse~C Rayan.
\newblock Understanding and mitigating bias in imaging artificial intelligence.
\newblock {\em Radiographics}, 44(5):e230067, 2024.

\bibitem{thiyagarajan2024physiology}
Dhanalakshmi~K Thiyagarajan, Hajira Basit, and Rebecca Jeanmonod.
\newblock Physiology, menstrual cycle.
\newblock In {\em StatPearls [Internet]}. StatPearls Publishing, 2024.

\bibitem{valdez2019circadian}
Pablo Valdez.
\newblock Circadian rhythms in attention.
\newblock {\em The Yale journal of biology and medicine}, 92(1):81, 2019.

\bibitem{vedhara2000acute}
Kavita Vedhara, J~Hyde, ID~Gilchrist, Michelle Tytherleigh, and Sue Plummer.
\newblock Acute stress, memory, attention and cortisol.
\newblock {\em Psychoneuroendocrinology}, 25(6):535--549, 2000.

\bibitem{vlasak2022neurocognitive}
Thomas Vlasak, Tanja Dujlovic, and Alfred Barth.
\newblock Neurocognitive impairment in night and shift workers: a meta-analysis of observational studies.
\newblock {\em Occupational and environmental medicine}, 79(6):365--372, 2022.

\bibitem{Wajcman01062007}
Judy Wajcman.
\newblock From women and technology to gendered technoscience.
\newblock {\em Information, Communication \& Society}, 10(3):287--298, 2007.

\bibitem{xavier2025biases}
Bibin Xavier.
\newblock Biases within ai: challenging the illusion of neutrality.
\newblock {\em AI \& SOCIETY}, 40(3):1545--1546, 2025.

\bibitem{xu2021relationship}
Shiyang Xu, Miriam Akioma, and Zhen Yuan.
\newblock Relationship between circadian rhythm and brain cognitive functions.
\newblock {\em Frontiers of optoelectronics}, 14(3):278--287, 2021.

\bibitem{xu2019emotionally}
Xiaobo Xu, Wenling Liu, and Weiguo Pang.
\newblock Are emotionally intelligent people more creative? a meta-analysis of the emotional intelligence--creativity link.
\newblock {\em Sustainability}, 11(21):6123, 2019.

\bibitem{zellers2019hellaswag}
Rowan Zellers, Ari Holtzman, Yonatan Bisk, Ali Farhadi, and Yejin Choi.
\newblock Hellaswag: Can a machine really finish your sentence?
\newblock In {\em Proceedings of the 57th Annual Meeting of the Association for Computational Linguistics}, 2019.

\end{thebibliography}

\end{document}